\documentclass{vgtc}                          

\ifpdf
  \pdfoutput=1\relax                   
  \usepackage{graphicx}                
  \DeclareGraphicsExtensions{.pdf,.png,.jpg,.jpeg} 
\else
  \usepackage{graphicx}                
  \DeclareGraphicsExtensions{.eps}     
\fi%

\graphicspath{{figures/}{pictures/}{images/}{./}} 

\usepackage{microtype}                 
\PassOptionsToPackage{warn}{textcomp}  
\usepackage{textcomp}                  
\usepackage{mathptmx}                  
\usepackage{times}                     
\usepackage{cite}                      
\usepackage{tabu}                      
\usepackage{booktabs}                  

\usepackage[normalem]{ulem}

\onlineid{0}

\vgtccategory{Research}

\vgtcinsertpkg

\newcommand{\rev}[1]{\textcolor{blue}{#1}}
\newcommand{\revtwo}[1]{\textcolor{blue}{#1}}
\newcommand{\off}[1]{\textcolor{red}{#1}}
\newcommand{\offtwo}[1]{\textcolor{red}{#1}}

\renewcommand{\rev}[1]{#1}
\renewcommand{\revtwo}[1]{#1}
\renewcommand{\off}[1]{}
\renewcommand{\offtwo}[1]{}

\title{The Impact of Virtual Reality and Viewpoints\\in \off{Wearable Telerobotics}\rev{Body Motion Based Drone Teleoperation}}

\author{
Matteo Macchini, \textit{Student Member, IEEE}, Manana Lortkipanidze, Fabrizio Schiano, \textit{Member, IEEE}, \\ and Dario Floreano, \textit{Senior Member, IEEE}
\thanks{The authors are with the Laboratory of Intelligent Systems, École Polytechnique Fédérale de Lausanne, CH-1015 Lausanne (EPFL), Switzerland.}%
}

\teaser{
 \includegraphics[width=\textwidth]{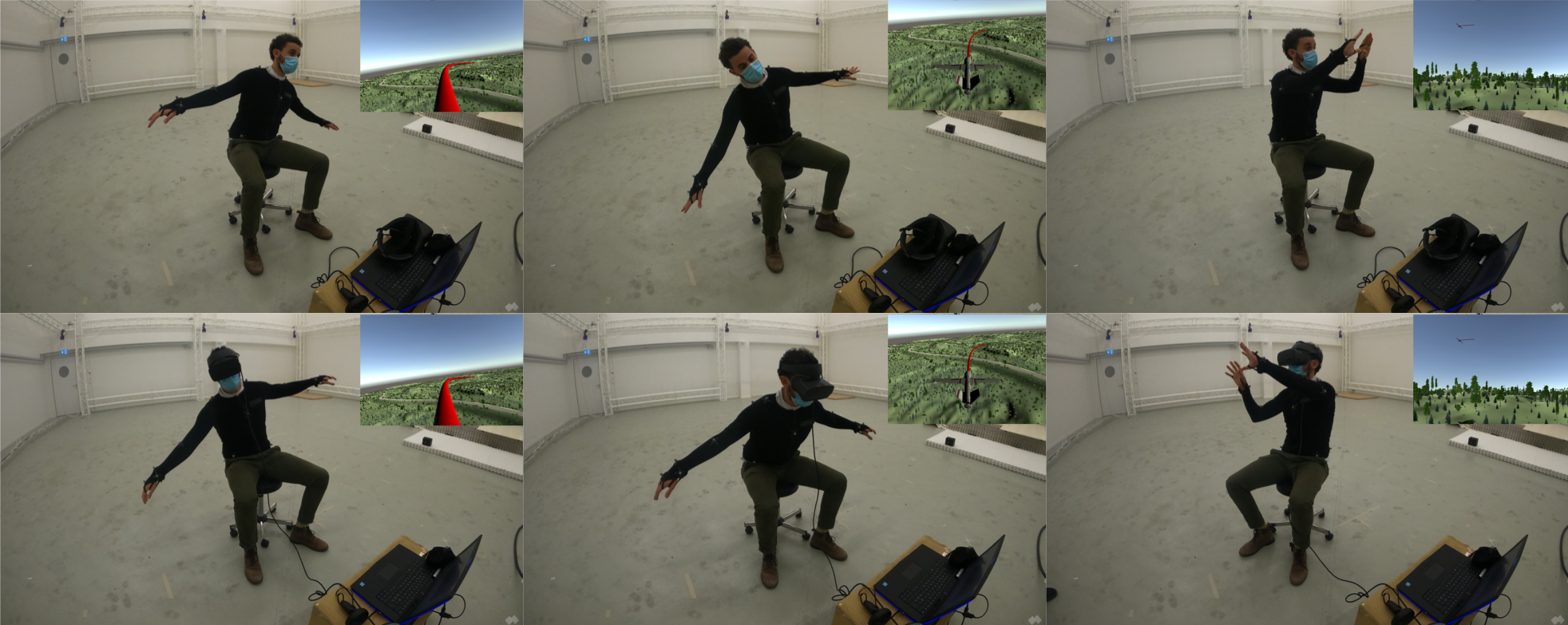}
 \caption{Spontaneous body motion acquisition during a robot imitation task using different viewpoints. The acquired data can be used to define personalized Body-Machine Interfaces. (top) VR-disabled conditions. (bottom) VR-enabled conditions.}
 \label{f:teaser}
}

\abstract{The operation of telerobotic systems can be a challenging task, requiring intuitive and efficient interfaces to enable inexperienced users to attain a high level of proficiency. 
Body-Machine Interfaces (BoMI) represent a promising alternative to standard control devices, such as joysticks, because they leverage intuitive body motion and gestures. 
It has been shown that the use of Virtual Reality (VR) and first-person view perspectives can increase the user’s sense of presence in avatars. 
However, it is unclear if these beneficial effects occur also in the teleoperation of non-anthropomorphic robots that display motion patterns different from those of humans. 
Here we describe experimental results on teleoperation of a non-anthropomorphic drone showing that VR correlates with a higher sense of spatial presence, whereas viewpoints moving coherently with the robot are associated with a higher sense of embodiment. 
Furthermore, the experimental results show that spontaneous body motion patterns are affected by VR and viewpoint conditions in terms of variability, amplitude, and robot correlates, suggesting that the design of BoMIs \offtwo{for robotic teleoperation} \revtwo{for drone teleoperation} must take into account the use of Virtual Reality and the choice of the viewpoint.

\vspace{2mm}
}

\CCScatlist{
  \CCScatTwelve{Human-centered computing}{Human computer interaction (HCI)}{Interaction paradigms}{Virtual reality}
  \CCScatTwelve{Human-centered computing}{Human computer interaction (HCI)}{HCI design and evaluation methods}{User studies}
}

\newcommand{\reffig}[1]{Fig. \ref{#1}}

\usepackage{colortbl}

\begin{document}


\maketitle



\section{Introduction}

Telerobotic systems are needed in many fields in which human cognition and decision-making capacities are still crucial to accomplish a mission \cite{gibo_shared_2016}.
Such fields include but are not limited to navigation in challenging and unstructured environments, search and rescue missions, and minimally invasive surgery \cite{diftler_robonaut_2011, khatib_ocean_2016, murphy_search_2008, bodner_first_2004}.
To provide fine control of the telerobotic system, the implementation of an efficient Human-Robot Interface (HRI) is crucial. 
Most telerobotic applications are currently restricted to a small set of experts who need to undergo long training processes to gain experience and expertise in the task \cite{chen_human_2007, casper_human-robot_2003}. 
With the fast advancements in the field of robotics, new systems require control interfaces that are sufficiently powerful and intuitive also for inexperienced users \cite{peschel_humanmachine_2013}.

Body-Machine Interfaces (BoMIs) are the subdomain of HRIs that consist of the acquisition and processing of body signals for the generation of control inputs for the telerobotic system \cite{casadio_body-machine_2012}.
BoMIs are showing great potential in improving user's comfort and performance during the operation of mobile robots \cite{miehlbradt_data-driven_2018}, representing a more intuitive alternative to standard interfaces.
The acquisition of body motion is particularly suitable to be applied to BoMIs due to the natural control capabilities that humans exert and train on it by daily activities.
The scientific literature offers several examples of motion-based interfaces, proposing a heterogeneous set of robots to be controlled and a multitude of methods to track the operator’s motion.

Among other mobile robots, drones are showing disruptive potential in both industrial and research applications \cite{floreano_science_2015, sesar_european_2016,delafontaine2020}.
In the implementation of motion-based BoMIs, one of the most challenging aspects is the definition of a \textit{mapping} function, which translates the user's body motion into robot commands.
For non-anthropomorphic systems, mapping functions can be designed based on the observation of the user's spontaneous behavior while they imitate the motion of the robot performing a set of predefined actions (\reffig{f:teaser}).
This procedure, known as \textit{calibration} or \textit{imitation} phase, has been used in prior works to identify the spontaneous motion patterns of users for the control of different robots \cite{macchini_personalized_2020, miehlbradt_data-driven_2018, pierce_data-driven_2012}.
Some key metrics used to discriminate the relevance of body motion features are their motion amplitude and their correlation with the robot's motion.
Some studies have been dedicated solely to the identification of common human motion patterns following this paradigm, in the case of both anthropomorphic and non-anthropomorphic robotic systems \cite{pierce_data-driven_2012, cauchard_drone_2015}.
Several interfaces have been proposed for the motion-based control of drones, based on the use of different body parts such as hands, torso, or the user's full body \cite{macchini_hand-worn_2020, rognon_flyjacket:_2018, sanna_kinect-based_2013}.
These works show that BoMIs can outperform standard interfaces, such as remote controllers, both in terms of performance and in user's personal preference.

Among the metrics used to qualify an HRI, \textit{presence} is a subjective measure of the feeling of "being there" in the virtual or distal scene \cite{slater_depth_1994}. 
Different examples in the literature support the hypothesis that increasing the sense of presence of an operator can improve their performance in controlling the robot \cite{toet_toward_2020, ma_presence_2006, song_flight_noyear}.
The concept of presence can be split into three dimensions: \textit{spatial presence},  \textit{self-presence}, and \textit{social presence}\cite{lee_presence_2004}.
In our study, we focus on the first two as social presence requires the interaction with different agents, which is not the case in most telerobotic missions.
While spatial presence is relative to the feeling of being surrounded by the virtual environment, self-presence defines the shift of the user's perception of self from their own body into the virtual or distal one \cite{kilteni_sense_2012}. 
A strong sense of self-presence can improve the operator’s sense of embodiment and give them the sensation of being the robot, instead of merely controlling it. 
The sense of self-presence and the concept of embodiment are thus closely related. 
Several factors concur to enhancing the sense of presence in the user during teleoperation, including the insulation from the real environment, and the amount and type of provided multi-modal feedback \cite{sanchez-vives_maria_presence_2005}.

The change of viewpoint can strongly alter the perception of a virtual environment.
Some works correlate the first-person view (1V) to a higher sense of embodiment \cite{slater_first_2010, petkova_perspective_2011}.
Studies in the field of video games conclude that 1V is also associated with higher performance in manipulation tasks, and in general when an interaction with static objects is needed \cite{taylor_video_noyear}. 
The same research states that a third-person view (3V) can increase the user's sense of spatial presence in the environment, which translates into a higher capacity of navigation and perception of the surrounding space.
Nonetheless, this advantage comes at the cost of a lower sense of embodiment.
Gorisse et al. studied the effects of different viewpoints (1V and 3V) on the sense of presence and embodiment during the control of an anthropomorphic avatar \cite{gorisse_first-_2017}. 
Through a survey, they concluded that the viewpoint impact on spatial presence is very limited. 
However, the first-person view positively and significantly affects embodiment.

VR applications are supported by a Head-Mounted Display (HMD), a device that a user wears on their head, which provides stereoscopic vision and can track the user's head and eyes to provide them with the ability to explore the virtual environment (VE).
Prior literature investigated the effects of viewpoint in VEs as a comparison with standard flat displays, confirming the higher degree of immersion in  1V \cite{debarba_characterizing_2015}.
Also, some studies show that large freedom of motion positively influences the sense of presence \cite{sanchez-vives_maria_presence_2005, slater_influence_1998}.

Despite the extensive work conducted on the link between VR, viewpoint, and presence for the control of anthropomorphic robots and virtual avatars, few studies have been conducted in the control of non-anthropomorphic robots, which are the most common type of robotic systems. 
Research shows that humans can identify themselves with agents presenting  
visual aspects and kinematics different from the human body, provided that
they present human-like motion and move synchronously with the user  \cite{aymerich-franch_non-human_2017, aymerich-franch_can_noyear}.
The same concept has been extended to virtual supernumerary robotic arms \cite{takizawa_exploring_2019}.

In this paper, we study the effects of (a) different viewpoints and (b) the use of VR on the teleoperation of a fixed-wing drone, \revtwo{as an example of} a non-anthropomorphic robot with non-human motion  behaviors. 
Specifically, we conducted a set of experiments to assess: 
\begin{itemize}
    \item The effects of viewpoint and VR on the user's sense of spatial presence and embodiment, when they perceive the environment and move as a fixed-wing drone.
    \item The effects of viewpoint and VR on the user's spontaneous body motion when they are asked to mimic the drone's behavior with their body. We focus our analysis on three dimensions of body motion: variability, correlation with the robot's movements, and gesture amplitude.
\end{itemize}

\begin{figure*}[h!]
\includegraphics[width=\textwidth]{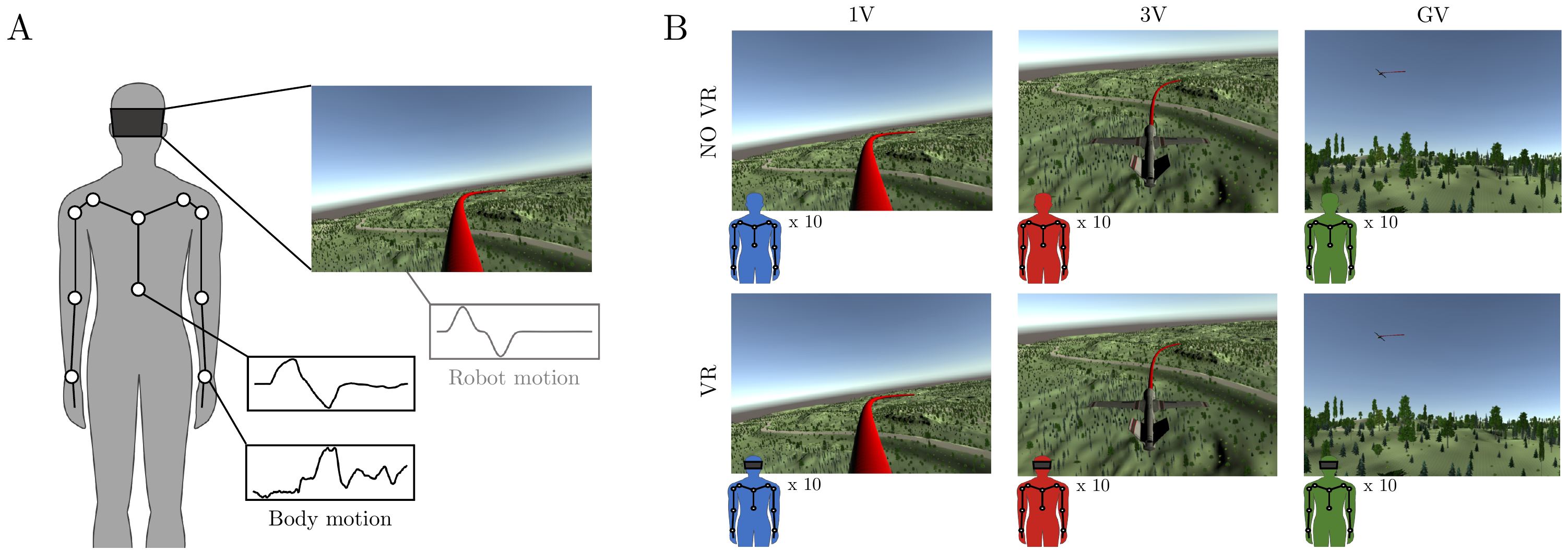}
	\caption{Overview of the experimental protocol. (A) Data acquisition scenario during the imitation task. The participant is free to move spontaneously to mimic the robot's behavior during a set of predefined maneuvers. During this phase, their body is tracked by a motion capture system, and synchronous data are acquired for the drone's trajectory and the user's motion. (B) Experimental conditions. A total of 30 participants took part in the experiment. We assigned each participant to one group, corresponding to one of the viewpoint conditions: first-person view (1V), third-person view (3V), or ground-view (GV). Each subject took part in the imitation task with and without the use of VR.}
	\label{f:protocol}
\end{figure*}

\section{Methods}

\subsection{Simulation}
The simulation environment used in this work is based on the Unity3D engine.
We used a robot model reproducing the dynamics of the commercial drone eBee by SenseFly\footnote{\href{https://www.sensefly.com}{https://www.sensefly.com}}.
The drone's attitude is stabilized through a PID controller and the speed is regulated to a constant value of $8m/s$.
The simulation displays a sequence of 4 maneuvers performed by the fixed-wing drone: two roll maneuvers (right, left), and two pitch maneuvers (up, down), after a horizontal flight section.
Each maneuver's duration was set to $T = 8s$ to give the participant enough time to recognize the robot's behavior, for a total duration of $32s$ per experiment. 
We display a red path in front of the drone to inform the participants of the drone's future trajectory (\reffig{f:protocol}B).
We considered 3 different conditions on the viewpoints: a standard first-person view (1V) from the drone's front camera, a third-person view from behind the drone (3V), and a third-person view from a ground observer (GV).
While most teleoperation tasks are carried out in 1V and GV condition, the inclusion of 3V allows us to decouple the effects of immersive viewpoints (1V vs. 3V, GV) and the effects of the camera moving together with the robot (1V, 3V vs. GV).
The 3V behind the drone follows at a constant distance and rotates with the drone.
The three viewpoints are depicted in \reffig{f:protocol}B.
We performed the experiments with two types of visual displays. 
In the VR condition, subjects used an Oculus Rift S HMD. 
In the non-VR condition, subjects saw the scene on a computer monitor positioned at a distance of $1.5m$.
In total, there are 6 experimental conditions, hereafter referred to as in Table \ref{t:conditions}.

\begin{table}[th]
\caption{Experimental conditions}
\renewcommand{\arraystretch}{1.1} 
\begin{center}
\begin{tabular}{  r  c  c }
 & \textbf{non-VR} & \textbf{VR} \\
 \hline
 \textbf{First-person view (1V)} & 1V-N & 1V-V \\
 \textbf{Third-person view (3V)} & 3V-N & 3V-V \\
 \textbf{Ground view (GV)} & GV-N & GV-V \\
 \hline
\end{tabular}
\label{t:conditions}
\end{center}
\end{table}

\subsection{Apparatus}
\label{ss:apparatus}
We run the experimental sessions in a room equipped with an OptiTrack Motion Capture (MoCap) System to track the participants' body.
Body motion tracking was performed through a set of 25 reflective markers strapped on a velcro vest worn by each subject.
The subjects' upper body was modeled as the concatenation of 13 different rigid bodies interconnected by sphere joints, as depicted in \reffig{f:protocol}A.
We consider only 9 rigid bodies for our study: torso, shoulders, arms, forearms, and hands.
\rev{This representation has already been adopted in relevant previous studies and demonstrated to be sufficiently powerful to derive personalized BoMIs for drone teleoperation \cite{miehlbradt_data-driven_2018, macchini_personalized_2020}.
Moreover, being the representation decoupled by construction, it prevents redundancy which might affect the subsequent data analysis.}
The orientation of each rigid body, expressed as a quaternion, was recorded at a frequency of $100$Hz.
During the experiments, we acquired synchronous data from both the drone simulator and the MoCap.
Encoded body pose and drone actions were streamed through a UDP protocol and concatenated into a dataset for subsequent analysis.

\subsection{Presence}
The sense of presence was measured through the post-experimental questionnaire shown in Table \ref{t:quest}. 
The questionnaire consisted of seven questions, where each item was given on a semantic scale from one to five (one corresponds to not at all, five corresponds to completely).
The questions were designed to investigate two dimensions of presence:
\begin{itemize}
    \item Embodiment, composed of three items representing two different embodiment dimensions: self-location and ownership of the virtual body \cite{kilteni_sense_2012}. The sense of agency was purposefully neglected since no actual teleoperation happens during our experiment. Questions are adapted from previous literature \cite{gorisse_first-_2017, debarba_characterizing_2015, argelaguet_role_2016}.
    \item Spatial presence, composed of four items, refers to the sense of environmental location and it is originally adapted from the MEC-SPQ test \cite{vorderer_mec_noyear}.
\end{itemize}

\begin{table}
\caption{Presence questionnaire composed of two blocks: Embodiment (E1-E3) and Spatial Presence (SP1-SP4).}
\renewcommand{\arraystretch}{1.1} 
\begin{center}
\begin{tabular}{  c p{7cm} } 
 \textbf{ID} & \textbf{Question} \\ 
 \hline
 E1 & To what extent did you feel that you were located inside the virtual body? \\ 
 E2 & To what extent did you feel that the virtual body was your own body? \\ 
 E3 & To what extent did you did you forget your actual body in favor of the virtual body? \\ 
 \hline
 SP1 & To what extent did you feel that you were actually there in the virtual environment? \\ 
 SP2 & To what extent did you feel that the objects in the virtual environment surrounded you? \\ 
 SP3 & To what extent did it seem to you that your true location had shifted into the virtual environment? \\
 SP4 & To what extent did you feel that you were physically present in the virtual environment? \\ 
  \hline
\end{tabular}
\label{t:quest}
\end{center}
\end{table}

\subsection{Participants}
30 volunteers participated in our user study.
All subjects had no know prior experience of motion sickness or discomfort using VR headsets, and a correct or corrected sight.
The age of the participants varied from 20 to 31 years ($24.29 \pm 2.78$) and 83\% of participants were male. 
Informed consent was obtained from everyone before the experiment and the study was conducted while adhering to standard ethical principles.\footnote{The experiments were approved by the Human Research Ethics Committee of the École Polytechnique Fédérale de Lausanne.}.

\subsection{Procedure}

Every subject filled a questionnaire about personal information before participating in the experiment.
Later, they were asked to sit on a stool in front of the computer and were shown the simulation through the screen or the VR headset depending on the group to which they were pseudo-randomly assigned.
During this procedure, they were asked to move spontaneously, as if they were controlling the drone's flight.
Since the maneuvers were predefined, no control was exerted on the simulator by the user.

Each participant took place in the calibration phase with two different conditions: with and without VR, and was pseudo-randomly assigned a viewpoint among the options above (1PV, 3V, 3PV). 
The simulation was shown twice per condition to each participant: a first one to get familiar with the scenario, and a second one for the actual data acquisition. 
The order of the experiment (with/without VR) they performed was determined pseudo-randomly to compensate for possible bias caused by the previous simulation. 
At the end of the experiment, each subject filled the presence questionnaire shown in Table \ref{t:quest}.

\subsection{Motion data preprocessing}

As mentioned in \ref{ss:apparatus}, raw data correspond to a set of signals consisting of time series of orientation information expressed in quaternions for the body motion, and two additional signals representing the drone roll and pitch angles during the experiment.
The user's upper body was modeled as a kinematic chain consisting of 9 rigid bodies, interconnected through sphere joints (\reffig{f:protocol}A).
We first computed the relative orientation of each rigid body with respect to its parent limb in the human kinematic chain.
For example, the shoulder rotation is expressed as the relative rotation with respect to the torso, and the arm rotation  with respect to the shoulder.
Subsequently, the initial rotation of the body segment was reset to zero to compensate for its initial bias and the orientations are converted from quaternions to Euler angles.
The order convention is chosen to minimize the risk of gimbal lock.
We filtered the Euler Angles with a moving average (N = 100) low-pass filter to mitigate the effects of quantization noise.
Finally, 27 angles (9 limbs x 3 angles) and 2 robot commands time series (roll and pitch) were used for our analysis.




\definecolor{LightYellow}{rgb}{1,1,0.9}
\definecolor{LightBlue}{rgb}{0.9,0.95,1}

\begin{table*}[t]
\renewcommand{\arraystretch}{1.1} 
\begin{center}
\caption{Mean and standard deviation of the responses to the presence survey in a scale 1 to 5}
\begin{tabular}{ c  c c c c c c } 
 \textbf{Condition} & \textbf{1V-N} & \textbf{3V-N} & \textbf{GV-N} & \textbf{1V-V} & \textbf{3V-V} & \textbf{GV-V} \\ 
\textbf{ID}  & \textbf{1N} & \textbf{3N} & \textbf{GN} & \textbf{1V} & \textbf{3V} & \textbf{GV} \\
\hline
E1  & $3.00 \pm 0.94$ & $2.92 \pm 1.19$ & $2.08 \pm 1.12$ & $3.60 \pm 0.84$ & $2.73 \pm 1.27$ & $1.82 \pm 1.08$ \\
E2  & $2.80 \pm 0.92$ & $2.92 \pm 1.04$ & $2.38 \pm 1.19$ & $3.50 \pm 1.08$ & $2.82 \pm 0.87$ & $1.82 \pm 0.98$ \\
E3  & $2.60 \pm 1.07$ & $2.62 \pm 1.12$ & $2.31 \pm 1.18$ & $3.40 \pm 1.07$ & $3.45 \pm 1.37$ & $2.09 \pm 1.04$ \\
\hline
SP1  & $2.40 \pm 0.84$ & $3.00 \pm 1.29$ & $2.85 \pm 1.14$ & $3.10 \pm 0.99$ & $4.00 \pm 1.26$ & $3.45 \pm 1.04$ \\
SP2  & $2.20 \pm 0.79$ & $2.31 \pm 1.03$ & $2.31 \pm 1.11$ & $3.30 \pm 0.67$ & $3.64 \pm 1.12$ & $3.27 \pm 1.01$ \\
SP3  & $2.20 \pm 0.79$ & $2.85 \pm 1.07$ & $2.62 \pm 1.12$ & $3.30 \pm 0.95$ & $3.91 \pm 1.22$ & $3.00 \pm 1.00$ \\
SP4  & $2.00 \pm 0.82$ & $2.69 \pm 1.18$ & $2.23 \pm 1.17$ & $3.10 \pm 0.74$ & $3.73 \pm 1.35$ & $3.09 \pm 1.38$ \\
\hline
\end{tabular}
\label{t:res1}
\end{center}
\end{table*}

\begin{table*}[t]
\renewcommand{\arraystretch}{1.1} 
\begin{center}
\caption{Statistical significance of the responses to the presence survey. $p_{1N,1V}$, $p_{3N,3V}$, $p_{GN,GV}$ refer to the VR condition, and are evaluated with Mann–Whitney U test. $p_{1N,3N}$,  $p_{1N,GN}$,  $p_{3N,GN}$,  $p_{1V,3V}$,  $p_{1V,GV}$,  $p_{3V,GV}$ refer to the viewpoint condition, for which we used the Wilcoxon Signed-Ranks test. Significant p-values are shown in bold characters. p-values are considered significant if $p<0.05$. In yellow and blue, the sections relative to the most relevant results.}
\begin{tabular}{c c c c}  

 \multicolumn{1}{c}{} & \multicolumn{3}{c}{}\\ 
 
 \multicolumn{1}{c}{} & \multicolumn{3}{c}{\textbf{VR Effect}} \\ 
 
 & \textbf{$p_{1N,1V}$} & \textbf{$p_{3N,3V}$} & \textbf{$p_{GN,GV}$}\\ 
 \hline
E1  & $0.084$ & $0.382$ & $0.297$\\
E2  & $0.089$ & $0.463$ & $0.106$\\
E3  & $0.052$ & $0.052$ & $0.357$\\
\hline
SP1  & \cellcolor{LightBlue}$0.064$ & \cellcolor{LightBlue}$\textbf{0.028}$ & \cellcolor{LightBlue}$0.133$\\
SP2  & \cellcolor{LightBlue}$\textbf{0.004}$ & \cellcolor{LightBlue}$\textbf{0.003}$ & \cellcolor{LightBlue}$\textbf{0.026}$\\
SP3  & \cellcolor{LightBlue}$\textbf{0.010}$ & \cellcolor{LightBlue}$\textbf{0.011}$ & \cellcolor{LightBlue}$0.199$\\
SP4  & \cellcolor{LightBlue}$\textbf{0.005}$ & \cellcolor{LightBlue}$\textbf{0.028}$ & \cellcolor{LightBlue}$0.065$\\
\hline
\end{tabular}
\hspace{10mm}   
\begin{tabular}{c c c c c c c}  
 \multicolumn{1}{c}{} & \multicolumn{6}{c}{\textbf{Viewpoint Effect}} \\ 
 \multicolumn{1}{c}{} & \multicolumn{3}{c}{\textbf{non-VR}} & \multicolumn{3}{c}{\textbf{VR}} \\ 
  & 
 \textbf{$p_{1N,3N}$} & \textbf{$p_{1N,GN}$} & \textbf{$p_{3N,GN}$} & \textbf{$p_{1V,3V}$} & \textbf{$p_{1V,GV}$} & \textbf{$p_{3V,GV}$} \\ 
 \hline
E1  & $0.411$ & $\textbf{0.028}$ & $\textbf{0.043}$ & \cellcolor{LightYellow}$0.064$ & \cellcolor{LightYellow}$\textbf{0.001}$ & \cellcolor{LightYellow}$\textbf{0.049}$\\
E3  & $0.447$ & $0.129$ & $0.087$ & \cellcolor{LightYellow}$0.085$ & \cellcolor{LightYellow}$\textbf{0.002}$ & \cellcolor{LightYellow}$\textbf{0.011}$\\
E5  & $0.487$ & $0.272$ & $0.244$ & \cellcolor{LightYellow}$0.341$ & \cellcolor{LightYellow}$\textbf{0.008}$ & \cellcolor{LightYellow}$\textbf{0.015}$\\
\hline
SP1  & $0.117$ & $0.151$ & $0.374$ & $\textbf{0.029}$ & $0.220$ & $0.092$\\
SP2  & $0.383$ & $0.436$ & $0.500$ & $0.128$ & $0.485$ & $0.176$\\
SP3  & $0.056$ & $0.209$ & $0.297$ & $0.067$ & $0.315$ & $\textbf{0.022}$\\
SP4  & $0.087$ & $0.397$ & $0.159$ & $0.072$ & $0.485$ & $0.133$\\
\hline
\end{tabular}
\label{t:res2}
\end{center}
\end{table*}

\section{Results}

In this section, we report on the experimental results regarding the presence survey, and the users' body motion analysis.

\subsection{Presence Survey}

As a preliminary validation, Cronbach’s alpha coefficient was calculated to measure the reliability of the questionnaire across both dimensions for all of the experimental conditions \cite{bland_statistics_1997}. 
The coefficient was $ > 0.73$ for all cases, thus indicating acceptable reliability of the questionnaire.
The Shapiro–Wilk test was carried out to check the normality of the distributions of the answers to the post-experiment questionnaire~\cite{shapiro_analysis_2020}.
As not all variables followed a normal distribution, we used a non-parametric test (the Wilcoxon Signed-Ranks) to compare objective performance data of paired groups, keeping the VR variable constant (1V vs 3V vs GV) \cite{wilcoxon_individual_1992}. 
We used the Mann–Whitney U test for the independent groups, keeping the viewpoint constant (VR vs non-VR) \cite{mann_test_1947}.

Table \ref{t:res1} shows the survey results across all conditions and Table \ref{t:res2} the statistical significance of our findings.
Each column refers to a pair of conditions, with $p_{a, b}$ representing the p-value relative to the t-test applied between condition $a$ and condition $b$.

\begin{figure}[t]
\begin{center}
\includegraphics[width=\columnwidth]{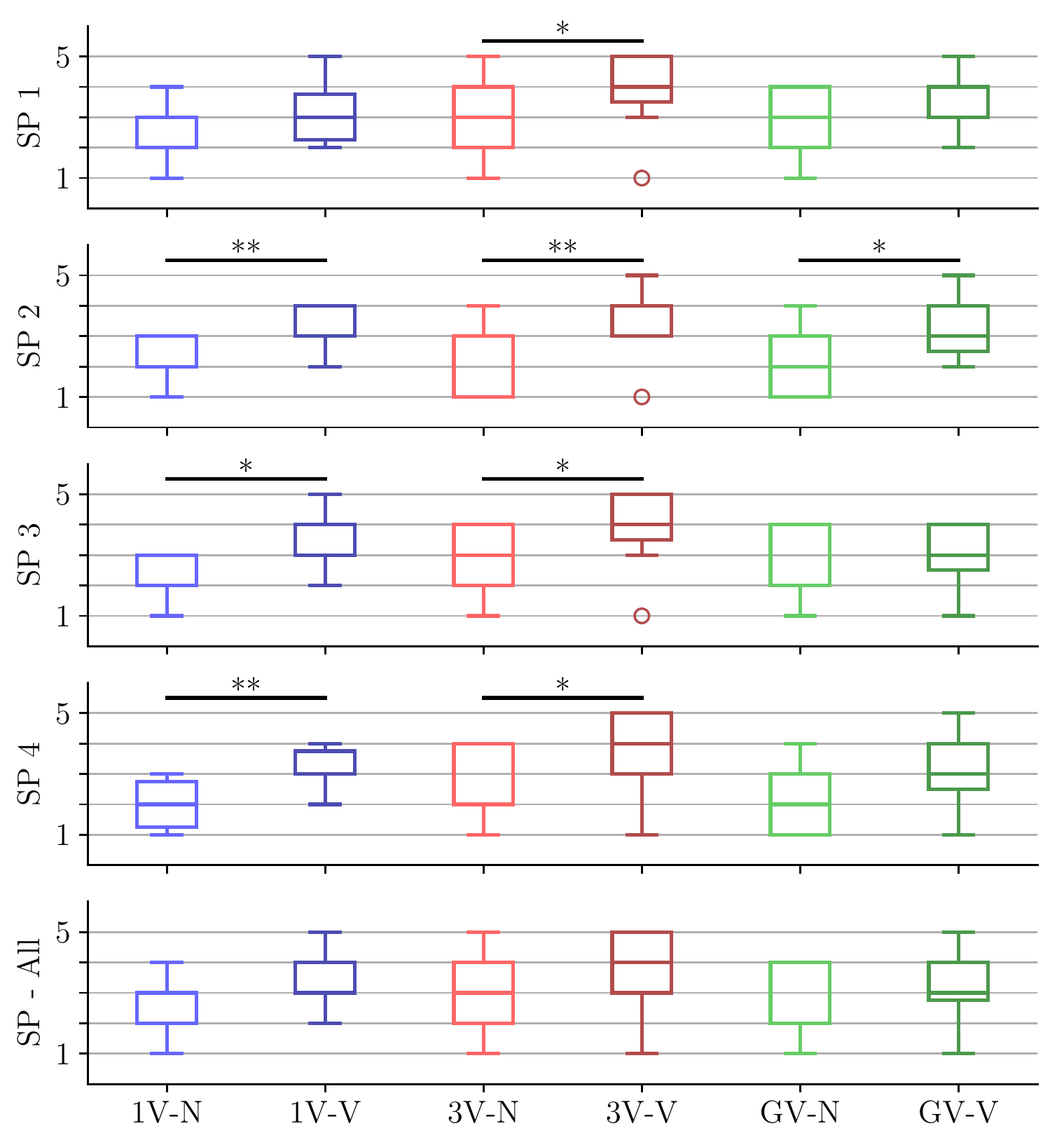}
	\caption{Survey results relative to the Spatial Presence dimension and average responses
	(Table \ref{t:res2}, blue highlight). 
	($^{**}p<0.01$, $^{*}p<0.05$)}
\label{f:surv_SP}
\end{center}
\end{figure}

\textbf{\textit{VR affects Spatial Presence.}}
In Table \ref{t:res2} (left) we show the survey results across the VR condition. 
The p-values are calculated for each viewpoint varying the VR condition using the Mann–Whitney U test.
From E1-E3 in Table \ref{t:res2} (left), we can observe that no significant difference was present between the VR groups and the non-VR groups ($p_{1N,1V},p_{3N,3V},p_{GN,GV}>0.05$). 
We cannot thus conclude that the use of VR affects the sense of embodiment.
Instead, SP1-SP4 show significant differences for almost all cases for both 1V and 3V.
The first-person view group reported higher values of spatial presence when using VR \rev{($SP_{1V-N} = 2.20 \pm 0.78$, $SP_{1V-V} = 3.20 \pm 0.81$).}\off{for SP2 ($SP2_{1V-N} = 2.40 \pm 0.84$, $SP2_{1V-V} = 3.10 \pm 0.99$, $p < 0.01$), SP3 ($SP3_{1V-N} = 2.20 \pm 0.79$, $SP3_{1V-V} = 3.30 \pm 0.95$, $p = 0.01$), and SP4 ($SP4_{1V-V} = 2.00 \pm 0.82$, $SP4_{1V-V} = 3.10 \pm 0.74$, $p < 0.010$).}
Also for third-person view the use of VR correlated with a higher sense of spatial presence \rev{($SP_{3V-N} = 2.71 \pm 1.13$, $SP_{3V-V} = 3.82 \pm 1.19$).}\off{for SP1 ($SP1_{3V-N} = 3.00 \pm 1.29$, $SP1_{3V-V} = 4.00 \pm 1.26$, $p = 0.028$), SP2 ($SP2_{3V-N} = 2.31 \pm 1.03$, $SP2_{3V-V} = 3.64 \pm 1.12$, $p < 0.01$), SP3 ($SP3_{3V-N} = 2.85 \pm 1.07$, $SP3_{3V-V} = 3.91 \pm 1.22$, $p = 0.011$), and SP4 ($SP4_{3V-N} = 2.69 \pm 1.18$, $SP4_{3V-V} = 3.73 \pm 1.35$, $p < 0.028$). }
This result suggests a higher sense of spatial presence for 1V-V compared to 1V-N and of 3V-V compared to 3V-N.
In average, the sense of spatial presence increased by $45\%$ for 1V and $40\%$ for 3V when using VR.
No significant effects were observed for the GV case.
These results are highlighted in light blue in Table \ref{t:res2} (left) and summarized in \reffig{f:surv_SP}.


\begin{figure}[h!]
\begin{center}
\includegraphics[width=\columnwidth]{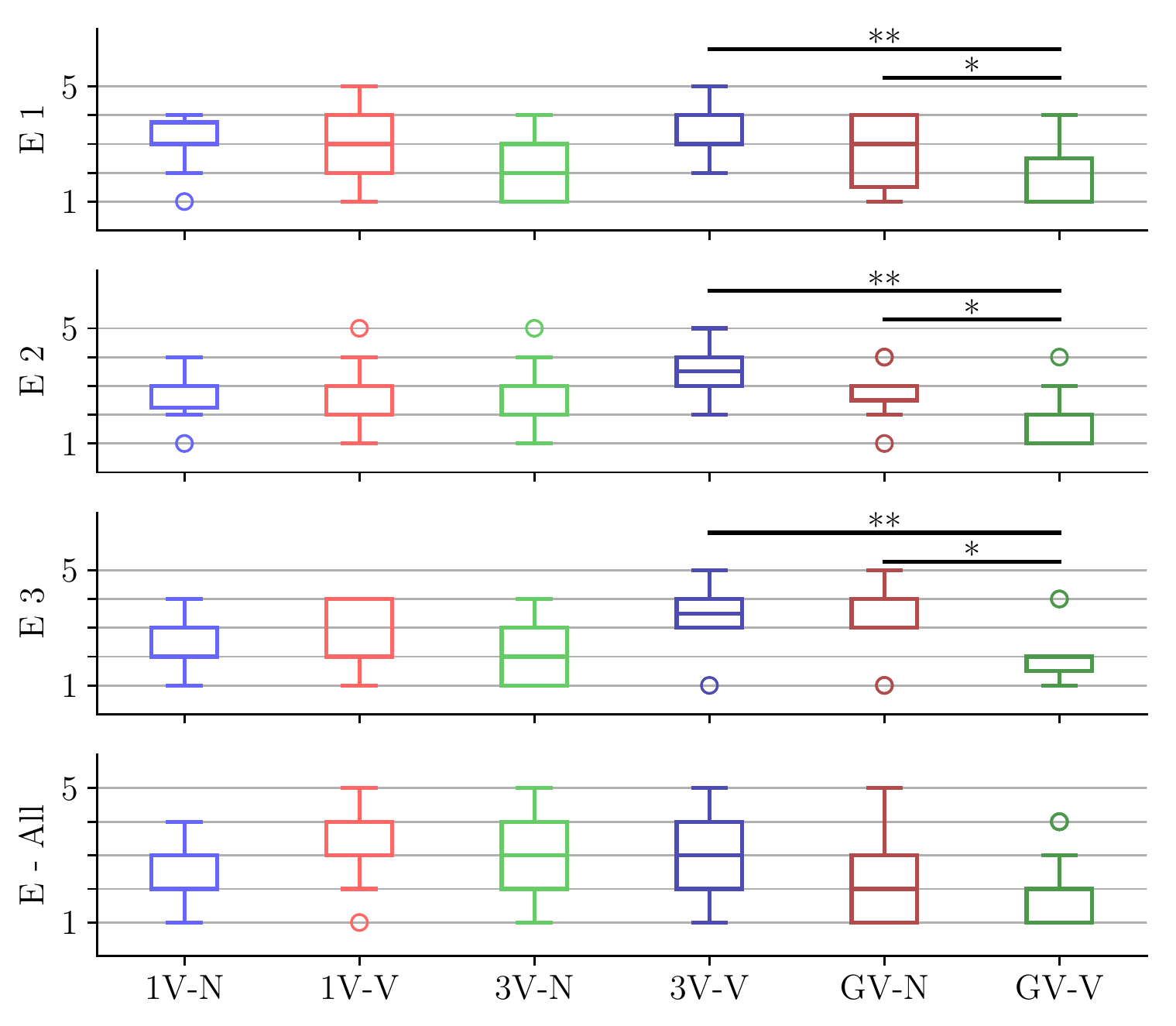}
	\caption{Survey results relative to the Embodiment dimension and average responses
	(Table \ref{t:res2}, yellow highlight). 
	($^{**}p<0.01$, $^{*}p<0.05$)}
\label{f:surv_E}
\end{center}
\end{figure}

\begin{figure*}[t]
\begin{center}
\includegraphics[width=\textwidth]{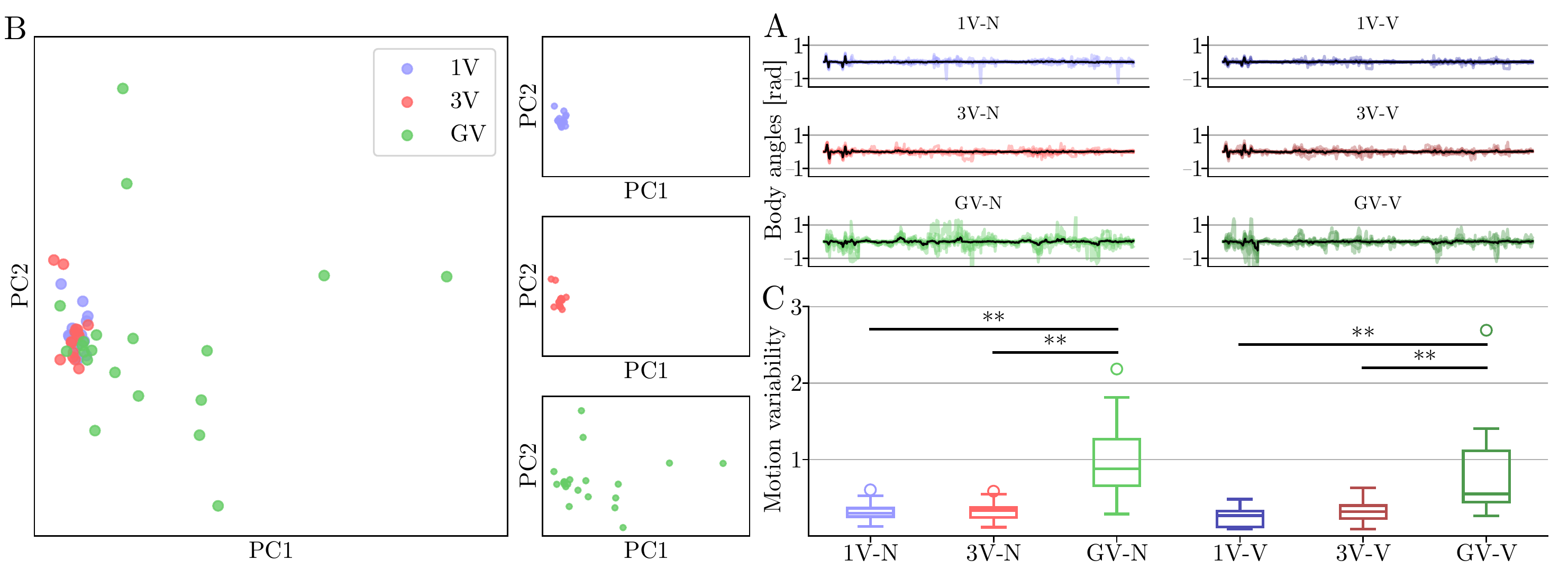}
	\caption{Observed motion variability during the imitation task.
	(A) Concatenated motion data timeseries. In black, the median value of the corresponding angle.
	(B) Scatter plot of the results of a two-dimensional PCA compression on the whole motion timeseries. Group GV exhibits a higher intra-group variability, while data belonging to groups 1V and 3V are more densely aggregated. 
	(C) Motion variability metric expressed as MSE from the median value. Both GV-N and GV-V values are significantly higher compared to the different viewpoints. ($^{**}p<0.01$)}
	\label{f:variab}
\end{center}
\end{figure*}

\textbf{\textit{Viewpoint affects Embodiment.}}
In Table \ref{t:res2} (right) we show the survey results across the viewpoint conditions.
The p-values are calculated between each pair of viewpoints using the Wilcoxon Signed-Ranks Test.
As a first observation, we can see that no significant effects are observable for any of the questions SP1-SP4, meaning that we cannot observe a significant correlation between the viewpoint and spatial presence.
Secondly, no significant effects are observable in the non-VR case.
On the other hand, there is a clear trend in the embodiment perception when changing the viewpoint while using VR: both $p_{1V,GV}$ and $p_{3V,GV}$ are significant in E1-E3.
GV group reported a lower sense of embodiment \rev{in average: ($E_{GV-V} = 1.91 \pm 1.00$) than 1V group ($E_{1V-V} = 3.50 \pm 0.96$, $p < 0.01$) and 3V group ($E_{3V-V} =3.00 \pm 1.18$, $p < 0.01$).} \off{for E1 ($E1_{GV-V} = 1.82 \pm 1.08$) than 1V group ($E1_{1V-V} = 3.60 \pm 0.84$, $p < 0.01$) and 3V group ($E1_{3V-V} =2.73 \pm 1.27$, $p = 0.049$).
Similar results were observed for E2 ($E2_{GV-V} =1.82 \pm 0.98$, $E2_{1V-V} =3.50 \pm 1.08$, $p < 0.01$, $E2_{3V-V} =2.82 \pm 0.87$, $p = 0.011$)
and E3 ($E3_{GV-V} =2.09 \pm 1.04$, $E3_{1V-V} =3.40 \pm 1.07$, $p < 0.01$, $E3_{3V-V} =3.45 \pm 1.37$, $p = 0.015$).}
The sense of embodiment was higher by $83\%$ for 1V and $57\%$ for 3V with respect to GV, only when using VR.
These results are highlighted in light yellow in Table \ref{t:res2} (right) and summarized in \reffig{f:surv_E}.

\subsection{Body Motion Analysis}

The second part of the data analysis was dedicated to the body motion data. These data were analyzed to extract relevant analogies and differences between the spontaneous body motion patterns during the calibration phase.
Due to the limited amount of subjects per condition (N=10), we chose non-parametric methods to assess the significance of our results.
For motion data, we used the Kruskal-Wallis test to assess the equality of the medians of different groups \cite{kruskal_use_1952}.
We focus on three aspects of motion, which are relevant for \offtwo{wearable}\revtwo{motion-based} teleoperation: motion variability, human-robot motion correlation, and gesture amplitude of different body segments.

\textbf{\textit{Spontaneous motion displays higher variability in Ground View.}}
First, we analyzed the intra-group motion variability. 
We concatenated motion data for each subject to a single timeseries containing the motion of all body segments (\reffig{f:variab}A).
To quantify the motion difference between subjects, the datasets were compressed to two-dimensional data using Principal Component Analysis (PCA) decomposition (\reffig{f:variab}B).
PCA emphasizes the data covariance across principal axes, revealing a main cluster of motion behaviors containing most of the 1V and 3V participants.
Oppositely, GV participant's motion is scattered further from the cluster center. 
Considering the centroid of the cluster formed by 1V and 3V participants, the average euclidean distance of the distribution is lower for 1V ($d_{1V} =0.83 \pm 0.60$) and 3V ($d_{3V} =1.17 \pm 1.16$) than for GV ($d_{GV} =7.00 \pm 6.93$, $p_{1V, GV}, p_{3V, GV} < 0.01$).

We then computed the median value of the concatenated features (\reffig{f:variab}B).
We computed the average MSE across all subjects in each group as a measure of intra-group motion variability (\reffig{f:variab}C).
Our results confirm the aforementioned observation: participants in GV condition moved more differently from each other, while groups 1V and 3V show a significantly lower motion variability, and thus a higher agreement with each other. 
Specifically, the variability of group GV-N ($variab_{GV-N} =1.01 \pm 0.46$) was higher than the variability of groups 1V-N ($variab_{1V-N} =0.30 \pm 0.11$, $p < 0.01$) and  3V-N ($variab_{3V-N} =0.33 \pm 0.13$, $p < 0.01$).
Similar results were observed comparing  GV-V ($variab_{GV-V} =0.78 \pm 0.52$) with 1V-V ($variab_{1V-V} =0.25 \pm 0.12$, $p < 0.01$) and 3V-V ($variab_{3V-V} =0.33 \pm 0.14$, $p < 0.01$).

\begin{figure}[h]
\begin{center}
\includegraphics[width=\columnwidth]{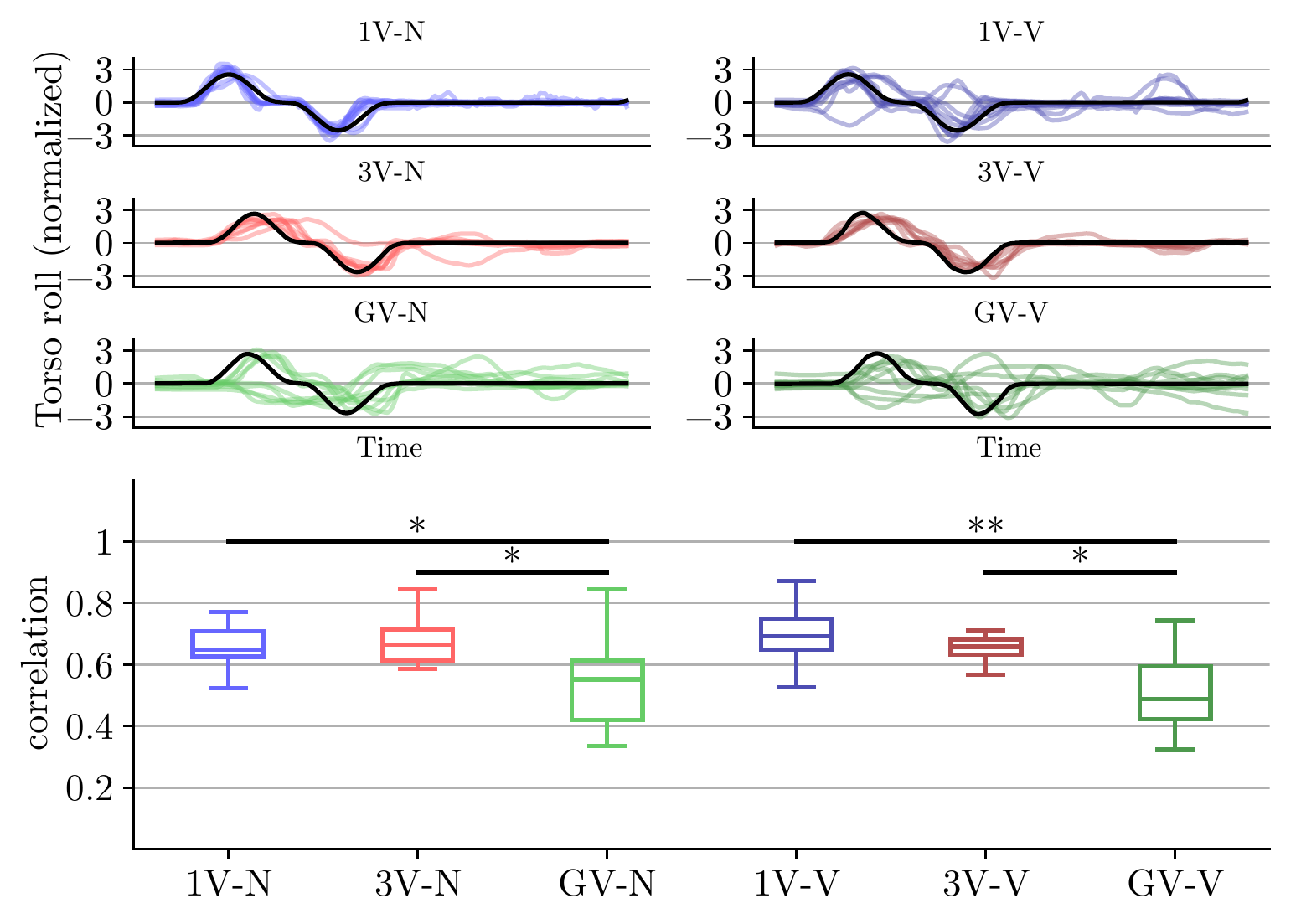}
	\caption{Correlation of human body motion and drone motion. \\(A) Drone roll superimposed to torso roll of different subjects. \\(B) Correlation score between human and robot motion. Both GV-N and GV-V conditions correspond to a lower correlation score. ($^{**}p<0.01$, $^{*}p<0.05$)}
	\label{f:corr_all}
\end{center}
\end{figure}

\textit{\textbf{Human-robot motion correlation is higher in First-Person View and Third-Person View.}}
Secondly, we considered the correlation between the user's movements and the drone's inputs, as a measure of similarity of the two datasets (\reffig{f:corr_all}).
We measured the correlation of a body segment with a robot command through the Pearson's correlation coefficient on the sample distribution.
We define the correlation of the Euler Angle related to a body segment with the robot motion as the sum of its absolute correlations with the drone roll and pitch. 
Additionally, we define the correlation of a body segment with the robot motion the maximum correlation value among all the associated Euler Angles.
Finally, we compute the mean across all body segments to evaluate the correlation score of the whole dataset.
We observed that group GV-N showed a lower correlation ($corr_{GV-N} =0.54 \pm 0.15$) than 1V-N  ($corr_{1V-N} =0.66 \pm 0.07$, $p = 0.034$) and 3V-N  ($corr_{3V-N} =0.68 \pm 0.08$, $p = 0.033$).
Similarly, group GV-V showed a lower correlation ($corr_{GV-V} =0.52 \pm 0.13$) than 1V-V  ($corr_{1V-V} =0.70 \pm 0.10$, $p < 0.01$) and 3V-V  ($corr_{3V-V} =0.65 \pm 0.04$, $p = 0.047$).


\begin{figure}[h]
\begin{center}
\includegraphics[width=\columnwidth]{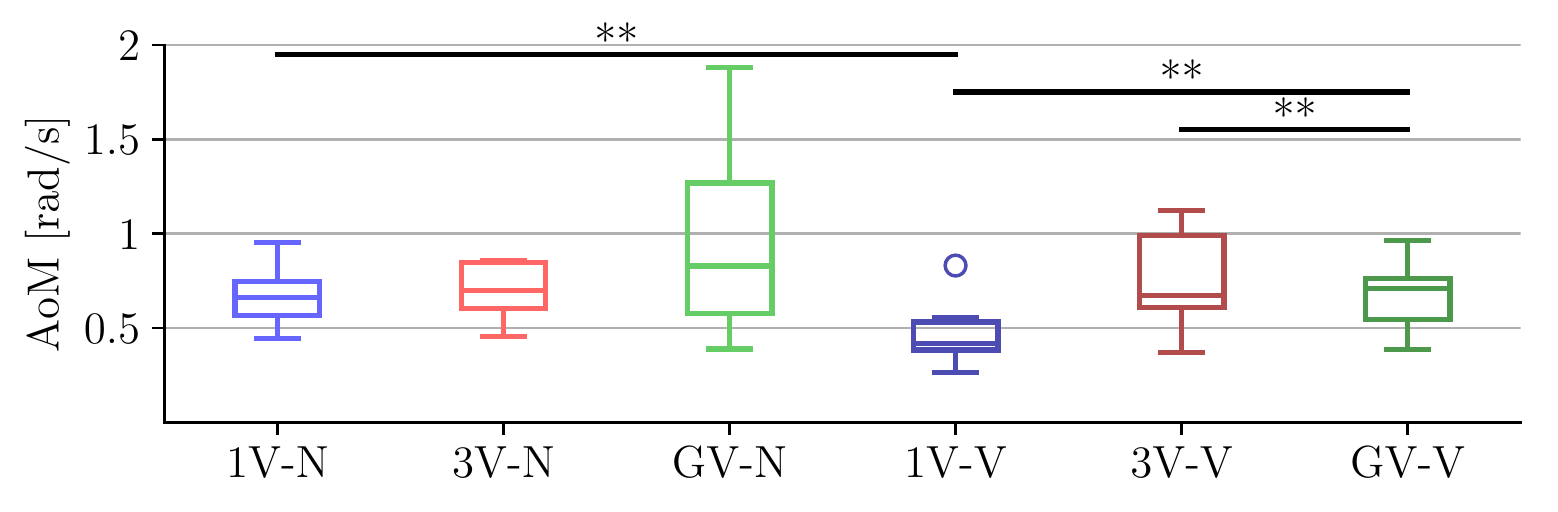}
	\caption{Amount of Motion score during the imitation task. 1V-V condition corresponds to smaller body gestures compared to both 1V-N and the other VR viewpoints, 3V-V and GV-V. ($^{**}p<0.01$, $^{*}p<0.05$)}
	\label{f:all_var}
\end{center}
\end{figure}

\textit{\textbf{VR and viewpoint affect motion amplitude in different ways.}}
Finally, we evaluated the differences in the gesture amplitude for the various body segments.
We define \textit{Amount of Motion} (AoM) of a body segment as the mean value of the norm of its angular velocity vector.
Our data indicate that the AoM significantly varies across conditions (\reffig{f:all_var}).
We observed two different effects of the viewpoint between VR and non-VR groups.
A first observation is that no significant VR effects are observable, except for group 1V-V. Group 1V-V employed smaller body gestures ($AoM_{1V-V} =0.46 \pm 0.15$) than group 1V-N ($AoM_{1V-N} =0.68 \pm 0.16$, $p < 0.01$).
Additionally, Group 1V-V moved significantly less than groups 3V-V  ($AoM_{3V-V} =0.77 \pm 0.26$, $p < 0.01$) and GV-V  ($AoM_{GV-V} =0.85 \pm 0.62$, $p < 0.01$), suggesting that users tend to move their torso more when they see the robot when controlling it in an immersive perspective.
Such an effect is not observable in VR-disabled experiments.

\section{Discussion}

In this study, we investigated how the use of VR and the viewpoint change affect two central aspects of \offtwo{wearable telerobotics}\revtwo{motion-based teleoperation}: the sense of presence and the user's spontaneous body motion\offtwo{for the operation of a}\revtwo{. We considered a fixed-wing} drone as an example of a non-anthropomorphic robot with non-human motion patterns. 
We run a user study (N=30) on an imitation task for motion pattern identification for \offtwo{wearable} \revtwo{motion-based} telerobotics systems (\reffig{f:protocol}), considering two experimental variables: the use of VR and the viewpoint (first-person: 1V; third-person: 3V;  ground: GV), for a total of six conditions.
Here, we discuss our main findings.


Our experiments produced two main sets of results. First, we observed that, for a given viewpoint, the use of VR can increase the sense of spatial presence.
VR-enabled experiments increased the user's sense of spatial presence for both 1V and 3V viewpoints up to $45\%$ (Table \ref{t:res1}, Table \ref{t:res2}, \reffig{f:surv_SP}). 
In comparison, the viewpoint perspective does not play such an important role on the sense of spatial presence.
However, VR significantly increases spatial presence only in 1V and 3V perspectives.

Furthermore, our data suggest that the viewpoint affects the user's sense of embodiment in VR-enabled experiments.
1V and 3V conditions provided a higher sense of embodiment (respectively, $83\%$ and $57\%$) compared with GV (Table \ref{t:res1}, Table \ref{t:res2}, \reffig{f:surv_E}). .
The same effect was not observed for VR-disabled experimental conditions.
These results are in agreement with prior literature~\cite{gorisse_first-_2017}, and show that they hold also for non-anthropomorphic robots with non-human motion patterns.

In summary, our study shows that the sense of presence in virtual environments is not exclusively limited to anthropomorphic characters with human-like motion, but can be elicited during the operation of other types of robots, such as the fixed-wing drone used here.
VR and viewpoint appear to affect different dimensions of teleoperation: while the first correlates with the sense of spatial presence, the second mainly impacts the sense of embodiment.
Moreover, the embodiment dimension seems to be related to a camera motion coherent with the robot' motion (1V, 3V) more than by using an immersive point of view (only 1V).
These effects do not apply to all conditions of VR and viewpoint: no change in the sense of spatial presence was observed in ground view, and no change in the sense of embodiment was observed without the use of VR.

Our second result concerns the human motion during a robot imitation task. We observed that human motion is mostly affected by the viewpoint perspective: while 1V and 3V groups presented similar intra-group motion patterns, subjects in group GV displayed much higher variability (\reffig{f:variab}) both in VR-enabled ($+223\%$) and VR-disabled conditions ($+169\%$). Since very different motion patterns cannot be recognized by a predefined BoMI, this result implies a higher need for personalization in applications where the teleoperation of the robot must take place with a ground view.

We also found a significant effect of the viewpoint on the correlation between the participants' and the robot's motion.
Specifically, condition GV reduces motion correlation up to $26\%$ in VR-enabled and by $35\%$ in VR-disabled experiments (\reffig{f:corr_all}).
This result can be explained by the higher difficulty to understand the robot's behavior in GV, since the drone is further away from the user and the perspective is not aligned with their view.
As human-robot motion correlation is a desirable feature for the definition of linear mapping functions, this result suggests that nonlinear mappings could be more effective for the definition of a BoMI for third-person view teleoperation.

Finally, gesture amplitude was affected in different ways by the viewpoint depending on the VR condition.
In VR-disabled experiments, between-groups motion amplitude did not change significantly.
Comparing with the VR-enabled condition, we observed that group 1V-V moved $32\%$ less than group 1V-N (\reffig{f:all_var}).
As it has been demonstrated that body motion is one of the main contributors to subjective presence in virtual environments \cite{slater_influence_1998}, it would be reasonable to expect that immersive experiences are associated with body movements with higher amplitude. 
Our results could be explained by the nature of our task: while previous literature focuses on active motion, in our study the subjects were passively following a trajectory, being able to rotate the camera but not to change the robot's motion.
Consequently, Group 1V saw only a camera moving, without the possibility to control its motion and this may have reduced the user's involvement in the task, and thus reduced the amplitude of the body movements.
Also, group 1V-V moved their body less than groups 3V-V and GV-V by $40\%$ and $46\%$, respectively.
This result suggests that having the robot in the user's field of view encourages users to move more than using an immersive viewpoint, possibly due to the aforementioned reasons.

In summary, we observed that viewpoint, and particularly the GV perspective translates in both a higher motion variability between subjects and in a lower user-robot motion correlation, suggesting that in this condition the implementation of a motion-based BoMI for the control of the drone would require both a high level of personalization and nonlinear mapping methods.
These results correlate with our findings regarding the sense of presence, and, particularly, the dimension of embodiment: according to our findings, the use of a GV viewpoint translates in both a higher sense of embodiment (in VR) and in a set of more various motion patterns among different participants. 
Moreover, GV also corresponds to a lower correlation between human and robot motion.
Further, we observed a lower motion amplitude linked to 1V-V, compared to both 1V-N, 3V-V and GV-V.
This effect might be related to the nature of our task.

Although our tests showed clear effects of VR and viewpoint over the sense of presence and the user's spontaneous body motion in teleoperation, several aspects should be investigated in future research.
First, the study of different robotic systems with different types of motion patterns would be needed to assess if our results extend to other platforms.\offtwo{Despite this limitation, we consider our method sufficiently general to be extended to different and more complex robotic morphologies, as a sufficient condition for its application is the identification of the set of DoF of the robot.} \revtwo{The extension to different and more complex robots is left for future work. This extension would add substantial value to our findings, and assess the transferability of the method.} Second, the study of active teleoperation tasks, in addition to our passive imitation studies, could help to understand whether that has an effect on the sense of presence and thus on body motion.
\rev{However, this step will require a significant advancement in state-of-art methods for the automatic definition of motion-based HRIs, as the implementation of HRIs from arbitrary body motion can be challenging on such variable motion datasets.}
Finally, our study took into account a limited set of motion variables consisting of body joint angles.
Although this choice is backed by relevant literature \cite{miehlbradt_data-driven_2018, rognon_flyjacket:_2018}, the observation of different kinematic variables (e.g., the position of the center of mass of the body segments) could unveil further results.

\section{Conclusions}

In this paper, we show new results explaining the effects of VR and viewpoint on the user's sense of presence and their spontaneous motion in teleoperation tasks.
Our findings provide new insights on presence for teleoperation in the case of a non-anthropomorphic robot with non-human motion, such as fixed-wing drones.
Additionally, we show that the users' motion patterns when mimicking the robot's motion are affected by VR and viewpoint conditions.
Our work suggests preferred experimental conditions for the definition of personalized body-machine interfaces \revtwo{for such machines}.
Since personalized interfaces have been demonstrated to be more effective than generic ones for the control of fixed-wing drones, the application of these results could facilitate the design and the implementation of new \offtwo{wearable,}motion-based telerobotic systems.

\section{Acknowledgements}

\rev{This work was partially funded by the European Union’s Horizon  2020  research  and  innovation  programme  under grant  agreement  ID:  871479  AERIAL-CORE, the Swiss National Science Foundation (SNSF) with grand number 200021-155907, and the National Centre of Competence in Research (NCCR) Robotics}

\bibliographystyle{abbrv-doi}

\bibliography{bib/alias,bib/IEEEConfAbrv,bib/IEEEabrv,bib/otherAbrv,bib/bibCustom}

\end{document}